\documentclass[sigconf,authorversion,nonacm]{acmart}

\usepackage{diagbox}
\usepackage{multirow}
\AtBeginDocument{%
  }

\begin{document}

\title{SD-GAN: Semantic Decomposition for Face Image Synthesis with Discrete Attribute}

%\begin{small}
\author{Kangneng Zhou}

\begin{small} \email{g20208857@xs.ustb.edu.cn}  \end{small}

\affiliation{%
\begin{small}  \institution{Department of Computer Science, University of Science and Technology Beijing, China}  \end{small}
\begin{small}  \country{}  \end{small}
}

\author{Xiaobin Zhu}
\begin{small}  \email{zhuxiaobin@ustb.edu.cn} \end{small}
\authornote{Corresponding author.}
\affiliation{%
\begin{small}   \institution{Department of Computer Science, University of Science and Technology Beijing, China} \end{small}
\begin{small}   \country{} \end{small}
}

\author{Daiheng Gao}
\begin{small} \email{daiheng.gdh@alibaba-inc.com} \end{small}
\affiliation{%
\begin{small}   \institution{DAMO Academy, Alibaba Group, China} \end{small}
\begin{small}   \country{} \end{small}
}

% \author{Kai Lee, Xinjie Li}
% \begin{small} \email{m13121032812@163.com} \email{abcdvzz@hotmail.com}  \end{small}
% \affiliation{%
% \begin{small}  \institution{Department of Computer Science, University of Science and Technology Beijing}  \end{small}
% \begin{small}  \country{}  \end{small}
% }

\author{Kai Lee}
\begin{small} \email{b20200344@xs.ustb.edu.cn}  \end{small}
\affiliation{%
\begin{small}  \institution{Department of Computer Science, University of Science and Technology Beijing, China}  \end{small}
\begin{small}  \country{}  \end{small}
}

\author{Xinjie Li}
\begin{small} \email{abcdvzz@hotmail.com}  \end{small}
\affiliation{%
\begin{small}  \institution{Department of Computer Science, University of Science and Technology Beijing, China}  \end{small}
\begin{small}  \country{}  \end{small}
}

\author{Xu-Cheng Yin}
\begin{small} \email{xuchengyin@ustb.edu.cn}  \end{small}
\affiliation{%
\begin{small}  \institution{Department of Computer Science, University of Science and Technology Beijing, China}  \end{small}
\begin{small}  \institution{USTB-EEasyTech Joint Lab of Artificial Intelligence,University of Science and Technology Beijing, China}  \end{small}
\begin{small}  \country{}  \end{small}
  }
%\end{small} 

% \affiliation{%

% \affiliation{%

% \affiliation{%

% \affiliation{%

% \affiliation{%

% \affiliation{%

% \affiliation{%

\renewcommand{\shortauthors}{Zhou et al.}

\begin{abstract}
Manipulating latent code in generative adversarial networks (GANs) for facial image synthesis mainly focuses on continuous attribute synthesis (e.g., age, pose and emotion), while discrete attribute synthesis (like face mask and eyeglasses) receives less attention. Directly applying existing works to facial discrete attributes may cause inaccurate results. In this work, we propose an innovative framework to tackle challenging facial discrete attribute synthesis via semantic decomposing, dubbed SD-GAN. To be concrete, we explicitly decompose the discrete attribute representation into two components, i.e. the semantic prior basis and offset latent representation. The semantic prior basis shows an initializing direction for manipulating face representation in the latent space. The offset latent presentation obtained by 3D-aware semantic fusion network is proposed to adjust prior basis. In addition, the fusion network integrates 3D embedding for better identity preservation and discrete attribute synthesis. The combination of prior basis and offset latent representation enable our method to synthesize photo-realistic face images with discrete attributes. Notably, we construct a large and valuable dataset MEGN (Face Mask and Eyeglasses images crawled from Google and Naver) for completing the lack of discrete attributes in the existing dataset. Extensive qualitative and quantitative experiments demonstrate the state-of-the-art performance of our method. Our code is available at an anonymous website: https://github.com/MontaEllis/SD-GAN.
\end{abstract}



\begin{teaserfigure}
% \begin{center}
    \includegraphics[width=0.9\linewidth]{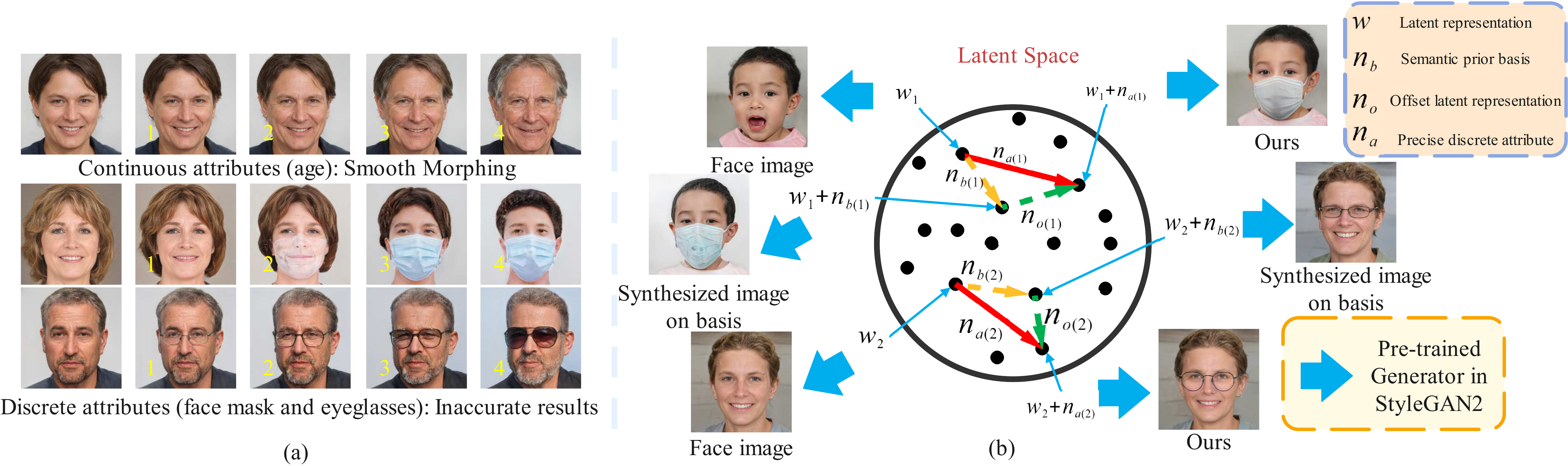}
% \end{center}
    \caption{Motivation and solution of our method. (a): Morphing results of continuous attributes (age) and discrete attributes (face mask and eyeglasses) via InterfaceGAN~\cite{interfacegan} with different length. Discrete attributes synthesis is inaccurate. (b): Schematic diagram of manipulating face representation in latent space. The yellow arrows indicate semantic prior basis $n_b$ for synthesizing images. The green arrows indicate the offset latent representation $n_o$ of facial attributes. The red arrows indicate the precise semantic representation $n_a$ in our method. $w$ is the face latent representation in the latent space of pre-trained StyleGAN2. Notice that each face representation has its unique discrete attribute code. Our method has the state-of-the-art performance.}
\label{fig:first}
\end{teaserfigure}


\maketitle

\section{Introduction}
Image synthesis has various applications, such as interactive graphics editing and image translation. With the rapid development of deep learning, image synthesis has achieved promising performance and received ever-increasing interest. Among different categories of natural images,  it is very challenging to synthesize discrete attributes of face images (e.g., face mask and eyeglasses) mainly due to the complicated structure of face images and the complex geometric relationships between face images and discrete attributes.

Image-to-image translation methods on face image synthesis try to learn mapping relationships among different image domains~\cite{cyclegan, pix2pix, pix2pixhq}. Generally, these methods achieve synthesis realism from appearance space while neglecting the critical geometry space. Some other techniques adopt image composition strategies to fuse foreground image (e.g., face mask) with background image (face image)~\cite{stgan,sfgan}. SF-GAN~\cite{sfgan} combines a geometry synthesizer with an appearance synthesizer to achieve synthesis realism.  Although these methods can keep other face attributes intact, they often suffer from the different distributions of two image domains, resulting in non coherent fusing edges.

Recently, learning facial semantics via manipulating latent code in the latent space has achieved great success in high-fidelity face image synthesis~\cite{interfacegan,ganspace,sefa}. GANSpace~\cite{ganspace} adopts PCA to find facial semantic representation in the latent space of the GAN model. StyleSpace~\cite{stylespace} utilizes style channels to control a highly disentangled visual attribute. These methods usually modify a latent code and enable semantic-level editing for generated images. They can synthesize faces with fine visual details, however, most of them mainly focus on continuous attributes (e.g., age, pose and emotion). When applying these methods to tackle facial discrete attribute synthesis, the results are always inaccurate. As shown in Fig.~\ref{fig:first}, the attribute representation in yellow arrows facilitates poor results which drift the accurate ones in red arrows. In addition, according to ~\cite{stylerig} the manifold corresponding to 2D images in the latent space do not allow to control accurately 3D shapes. The 3D-aware GANs have abundant 3D information but there are few works introduce them to facilitate 2D face image synthesis.

To address the above-mentioned problems, we propose an innovative framework (named SD-GAN) to decompose the latent code of facial and discrete attributes in the latent space of GANs. From our key observations (as shown in Fig.~\ref{fig:first}(a)), face image synthesis networks~\cite{interfacegan} fail in regressing accurate discrete attributes which are orthogonal to other facial attributes. For example, while manipulating face mask attribute on the second row of Fig.~\ref{fig:first}(a), the age and hair of the face changed as well. Hence, we decompose the semantic discrete attribute into prior basis and offset latent representation. As shown in Fig.~\ref{fig:first}, the optimal face images always correspond to different lengths of basis. So a novel search algorithm is also proposed for the optimal length of the basis. In this way, the semantic prior basis shows an initializing direction for manipulating face representation and makes the network focus on learning the following offset latent representation instead of losing its way in the large possible latent space.

In addition, highly motivated by the 3D controlling abilities of 3D-aware GAN~\cite{gram,eg3d,stylenerf,stylesdf,unsup3d, zhang2022multiview}, we propose a novel 3D-aware semantic fusion network to generate offset latent representation of discrete attributes for adjusting prior basis and performing better authentic. In this way, we introduce 3D embedding into 2D manifolds to help the network perform better in identity preservation and discrete attribute synthesis. The offset latent representation and semantic prior basis will be combined to facilitate the original latent code for generating the final synthesized image. As shown in Fig.~\ref{fig:first}, with the help ofW prior basis, the network easily converges to precise latent representation. Our method has the advantages of both face attribute intacting and visual details. Notably, we construct a dataset MEGN (Face Mask and EyeGlasses images crawled from Google and Naver). According to the experimental results in Sec.~\ref{exp:1},our MEGN greatly benefits the task of synthesizing images with discrete attributes. In summary, our contributions are:
\begin{itemize}
	\item We propose an innovative framework (named SD-GAN), decomposing semantic discrete attribute representation in the latent space of GANs into semantic prior basis and offset latent representation. Our method achieves state-of-the-art performance both qualitatively and quantitatively.

    \item We adopt the normal vector of the hyperplane created by a SVM classifier as the semantic prior basis and a novel optimal length of basis search algorithm is proposed therein. The semantic prior basis shows an initializing direction for manipulating face representation in the latent space. Also, a novel 3D-aware semantic fusion network is proposed to generate offset latent representation. 3D information is integrated in the network for achieving better authenticity.
    
    \item We propose a valuable dataset MEGN for complementing existing datasets that are lacking in discrete attributes, for the task of face image synthesis.
\end{itemize}

\begin{figure*}[!h]
\begin{center}
%   \includegraphics[width=1\linewidth]{img_new/refinenet_new_0723.eps}
   \includegraphics[width=0.85\linewidth]{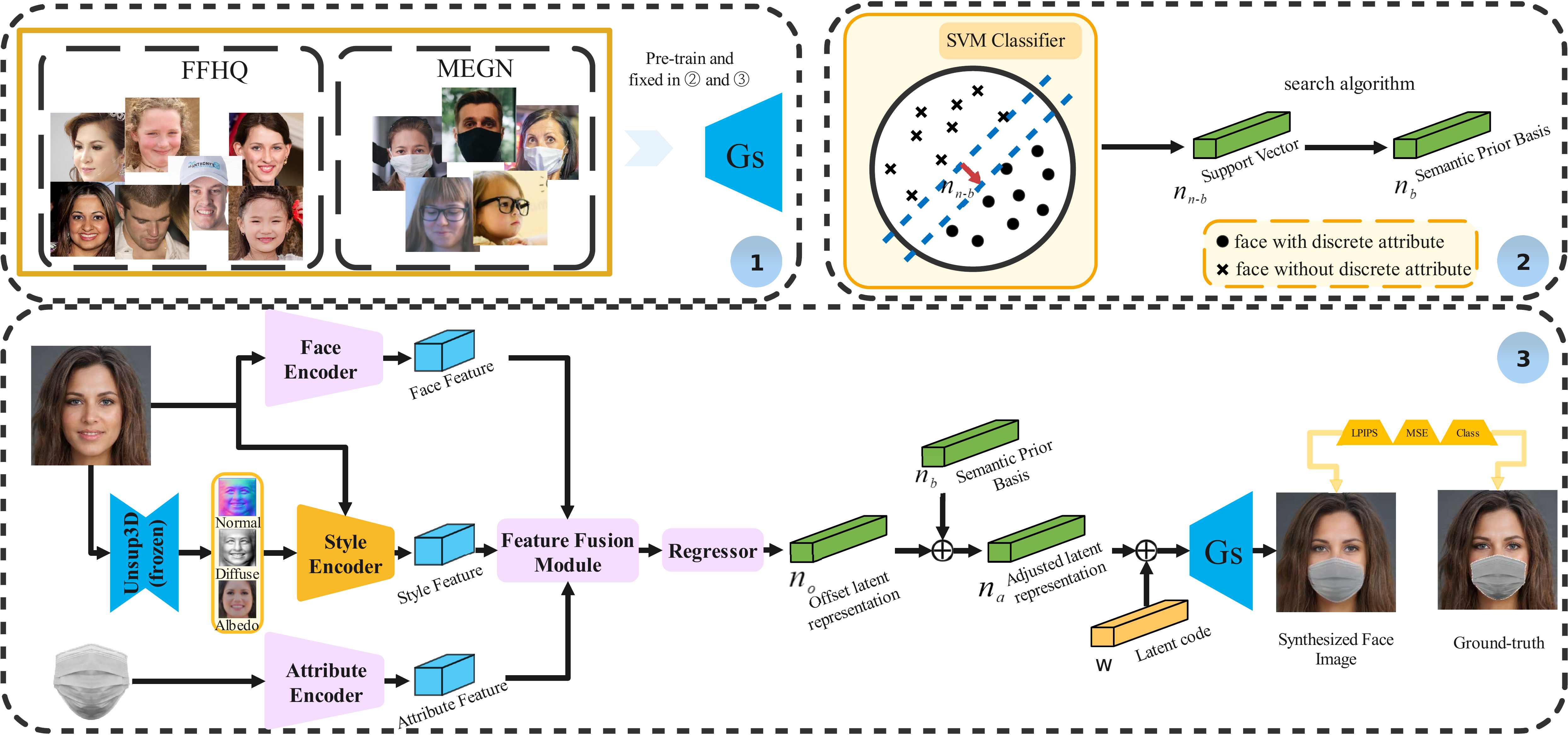}
\end{center}
   \caption{Framework of our method. (1): Diagram of StyleGAN2 pre-training. (2): Flowchart of generating semantic prior basis. (3): 3D-aware Fusion Network for offset latent representation. Our method decomposes facial attribute into semantic prior basis and offset latent representation, facilitating face synthesis with discrete attribute. The offset latent representation obtained by 3D-aware fusion network balances a better authenticity between face images and discrete attribute.}
\label{fig:refinenet}
\end{figure*}

\section{Related Work}
\subsection{Image Synthesis via GAN}
Image synthesis aims to fuse with other objects for generating realistic images while the remaining attributes in the un-covered part of the image are unchanged. Image-to-image translation methods try to learn optimal mapping relationships between different image domains for image synthesis. A lot of works adopt conditional GAN~\cite{pix2pix}, dual learning~\cite{pix2pix}, multi-domain transformation~\cite{cyclegan}, separated latent space swapping~\cite{stargan} and other novel methods~\cite{rsgan,paircyclegan,psgan} to synthesize general face attributes. Overall, the existing image-to-image translation methods often achieve synthesis realism from appearance space while neglecting the geometry space. Image-Composition based methods try to blend foreground images with a background image into a target image. ST-GAN~\cite{stgan} adopts geometric warping parameter space to synthesize images with geometric realism, but it neglects the appearance of realism. SF-GAN~\cite{sfgan} adopts a spatial fusion strategy to synthesize images for both appearance realism and geometric realism. 

Generally, image-composition based methods tend to suffer from different distributions of two image domains, resulting in incoherent fusing edges. Our method works in a semantic latent space rather than the methods mentioned above.

\subsection{Latent Semantic Manipulating}
Recently, image synthesis methods via manipulating the latent representation have achieved promising performance and attracted great attention. Chen \emph{et al.}~\cite{infogan} separates an input noise vector into an incompressible part and a latent code; hence it can concentrate on exploring semantic information underlying latent representation for generating synthesis images. StyleGAN~\cite{stylegan} is proposed to demonstrate the potential power of decoupling attributes in latent space to highlight the particular semantic attribute. 

Shen \emph{et al.}~\cite{interfacegan} hypothesize that the latent representations of two face images can be separated by a normal vector in latent space, which can be utilized to implement controllable face image synthesis. H{\"{a}}rk{\"{o}}nen \emph{et al.} \cite{ganspace} adopt PCA to find the principle face attribute representation in the latent space of GAN model. Shen and Zhou~\cite{sefa} propose an unsupervised method to find semantic representations for tackling some undefinable attributes, e.g., eye size and painting style. Tewari \emph{et al.}~\cite{stylerig} proposes a method to embed a parametric face model into the network and implement pose, illumination manipulation. StyleCLIP~\cite{styleclip} conducts image manipulation via a text-based interface for integrating the advantages of CLIP~\cite{clip} and StyleGAN. Hu \emph{et al.}~\cite{style_trans} investigates reference and label attribute editing through a pre-trained latent classifier. Overall, the existing GAN-based methods mainly focus on continuous attributes and often fail to hold discrete ones. Our method aims to synthesize faces with discrete attributes correctly.

\subsection{3D-aware GAN}
The works in GANs has promising performance, while the series of StyleGANs~\cite{stylegan, stylegan2, stylegan3} lack the ability to hold 3D controls and have difficulty to achieve complex editing. Recent works leverage Neural Radiance Field (NeRF~\cite{nerf}) to construct implicit fields to represent 3D scenes. The following works~\cite{pigan, stylenerf, cips3d, stylesdf} adopt periodic implicit GAN, progressively upsampling, Implicit Neural Representation~\cite{inr} and Signed Distance Function~(SDF)~\cite{deepsdf} to synthesize 3D-aware data. Head-NeRF~\cite{headnerf} is proposed to take 3DMM as prior to construct a face field. Other methods focus on wild images and reconstruct 3D faces in supervised~\cite{3ddfa, 3ddfav2} or unsupervised ways~\cite{unsup3d, gan2shape}. Our method embeds 3D information obtained by 3D-aware Unsup3d~\cite{unsup3d} into semantic fusion network to synthesize images with authenticity.

\section{Our Method}
\subsection{Overview}

The framework of our method is illustrated in Fig.~\ref{fig:refinenet}. In step one of Fig.~\ref{fig:refinenet}, we will first pre-train the generator of StyleGAN2 $G_s$ on FFHQ and our proposed MEGN. Then our generator will be fixed after pre-training. To decompose semantic attribute representation, we explore a semantic prior basis via an SVM classifier in the latent space in step two. 3D-aware semantic fusion network is proposed in step three, feature of face image and discrete attribute will be extracted by two individual encoders. The face image, and its corresponding 3D information extracted by Unsup3D~\cite{unsup3d}, will be combined to extract the feature by style encoder. The three features (face feature, attribute feature and style feature) will be used to learn offset latent representation by fusion module. The offset latent representation and semantic prior basis will be combined into an adjusted latent representation for promoting the latent representation of the original face image. The synthesized face image will be generated by the pre-trained generator $G_s$.

\subsection{Semantic Prior Basis}
The synthesis network of StyleGAN2~\cite{stylegan2} in the series of StyleGAN~\cite{stylegan,stylegan2,stylegan3} can be explained as a function $G_s$ that maps a latent code $w\in \mathbb{R}^{512}$ to a realistic face image $I=G_s(w)$. A hyperplane in the latent space serves as a decision boundary to separate the face attributes. Learning the hyperplane mainly consists of three steps. First, we sample $500k$ latent codes $w\in \mathbb{R}^{512}$ and generate the corresponding face images $I_s=G_s(w)$ using a pre-trained generator of StyleGAN2. Then, an attribute prediction model $F_{pred}$ will be adopted to compute a confidence score for the attribute of each image $conf = F_{pred}(I_s)$. We get the training set $\{w,conf\}$ and sort the corresponding scores and choose samples with extremely high scores as positive and extremely low ones as negative.

Finally, an SVM classifier will be trained among the dataset above and resulting in a decision boundary in the latent space. The normal vector of the decision boundary is normalized as semantic prior basis $n_{n-b}\in \mathbb{R}^{512}$ in the latent space of pre-trained StyleGAN2, as shown in Fig.~\ref{fig:refinenet}~(2).

% \begin{figure}[!t]
% \begin{center}
% %\fbox{\rule{0pt}{2in} \rule{0.9\linewidth}{0pt}}
%   \includegraphics[width=1\linewidth]{img_new/interfacegan.eps}
% \end{center}
%   \caption{Synthesis face attributes with InterfaceGAN. The red boxes represent the optimal images for wearing masks.}
% \label{fig:interfacegan}
% \end{figure}

However, the semantic prior basis obtained by the SVM classifier cannot distinguish different facial attributes (with or without discrete attribute) in the non-compact and unsmooth latent space of GAN. As shown in Fig.~\ref{fig:first}, the optimal images correspond to different lengths of prior basis. The face in the second row with length of 3 is the optimal while the face in the third row is optimal with weight of 2. In order to promote the initialized guide capability of prior basis, we adopt a novel search algorithm to compute an optimal length $\eta$ for the semantic prior basis of each face image. Specifically, we search for an optimal face image with discrete attributes while maintaining the identity as best as possible. We compute the length $\eta$ according to a formula that consists of three parts and formulated as below:

\begin{equation}
\label{eq:1}
\begin{split}
    Score= & F_{det}(G_{s}(w+\eta* n_{n-b}))  \\
    & + \lambda\times ||M\odot(G_{s}(w+\eta* n_{n-b})-G_{s}(w))||_2^2 \\
    & - \lambda\times ||\bar{M}\odot(G_{s}(w+\eta* n_{n-b})-G_{s}(w))||_2^2,
\end{split}
\end{equation}
\noindent where $F_{det}$ represents a discrete attribute detector which is trained by YOLO~\cite{yolo} whose confidence scores are used here, $G_s$ represents the synthesis network of generator in StyleGAN2, $w$ represents the latent code of face image ($w \in \mathbb{R}^{512}$), $M$ represents the binary mask of face which can be obtained by the method in Sec.~\ref{sec:synthesic}, and $\bar{M}$ represents the area which does not belong to binary face mask, $\lambda$ is used to balance loss terms and is set to $10$ here.

The first part in Eq.~\ref{eq:1} aims to force the network to learn the discrete attribute. The discrete attributes are harder than continuous ones to be disentangled among other attributes, so this part of loss is significant. The second part aims to morph the area of simulated discrete mask dramatically, which can improve the possibilities of correct manipulation and prevent the network from falling into local optima that do not update. The third part aim to maintain identity information while manipulating discrete attributes, which is significant for accurate discrete attribute manipulating.

We have a series of weights $\{\eta_1,\eta_2,...\}$ increasing linearly from $0$ to $10$ with step of 0.2 and selecting $\eta_m$ to maximize $Score$. The $\eta_m$ is regard as the optimal length and compute semantic prior basis:

\begin{equation}
    n_{b} = \eta_m *n_{n-b}
\end{equation}

% \begin{figure}[!htb]
% \begin{center}
% %\fbox{\rule{0pt}{2in} \rule{0.9\linewidth}{0pt}}
%   \includegraphics[width=1\linewidth]{img_new/basis_0723.eps}
% \end{center}
%   \caption{Flowchart of  generating semantic prior basis.}
% \label{fig:basis}
% \end{figure}

\subsection{3D-aware Semantic Fusion Network}
\label{sec:cfsnet}
Existing GAN-based methods generally adopt linear strategies to directly regress semantic representation~\cite{interfacegan}. These methods sometimes even fail in disentangling continuous attributes (e.g hair, smile, age), still less discrete attributes. Distinguished from existing methods, in our work, we aim to explicitly decompose the discrete attribute representation so that the network would focus on offset latent representation with the benefit of initializing direction of prior basis. The detailed structure of the 3D-aware semantic fusion network is illustrated in Fig.~\ref{fig:refinenet} (3). Given a face image $I_{f}$ generated by StyleGAN2~\cite{stylegan2} and a discrete attribute image $I_{m}$, the 3D-aware semantic fusion network learns a mapping function $f_m$ for correlating their latent semantic representations as:

\begin{equation}
    n_{o}=f_m(I_{f},I_{m}),
\end{equation}

\noindent where $n_{o}$ is the offset latent representation obtained by our semantic fusion network. It is worth noticing that $n_{o}\in \mathbb{R}^{14\times512}$ ($W+$ space which is more expressive than $W$ space). The $W+$ is an extended latent space which consolidates $14$ different $w\in\mathbb{R}^{512}$ code from $W$ space. The face latent codes $w$ here are sampled from the mapping network in StyleGAN2, so $w\in\mathbb{R}^{512}$. The $w$ can translate to $W+$ space via broadcasting. The Ground-truth of the training process can be found in Sec.~\ref{sec:synthesic}.

Here, we dissect the design of each sub-module (illustrated by pink boxes and yellow box in Fig.~\ref{fig:refinenet}). The face encoder is designed to learn object-specific features, which serves to preserve identity-information for the final results. Inspired by works in face recognition, the weights in face encoder are initialized by ArcFace~\cite{arcface} for better extracting unique face features at the beginning of training. For attribute encoder, we adopt the first four residual blocks of ResNet-18~\cite{resnet} to extract the feature. 

The fixed Unsup3d~\cite{unsup3d} is applied to get normal map, diffuse map and albedo. The normal map contains fine detailed shape of the face, while diffuse map and albedo represent the texture of the face. Our style encoder (the yellow box in Fig.~\ref{fig:refinenet}) consists of multi-layer convolutions to reflect sufficient 3D expression features from the concatenation of face normal map, diffuse map, and albedo images. In this way, the style encoder is expressive for 3D shape to estimate the position of discrete attributes and has adequate 3D information which can reconstruct the face to enhance the ability of identity preservation. With object-specific face feature, discrete attribute feature and adequate 3D feature, we propose a fusion module to fuse features and make them fully expressive for unique precise discrete attribute features. Inspired by SEAN~\cite{sean}, our feature fusion module adopts weighted learning with coefficients to fuse the extracted features. The details of network architecture of style encoder and fusion module can be found in Appendix.

Our regressor aims to map the fused feature to offset latent representation in the $W+$ space mentioned above. Our regressor consists of  sparse-connected layers~\cite{treeconnect} for mapping operation, which avoids the issue of redundant parameters in fully-connected layers. The structure not only has fewer parameters, but also has stronger performance capabilities.

So far, we get the adjusted latent code  $n_{a}$ through the addition of offset and $n_{b}$:

\begin{equation}
    n_{a}=n_{o} + n_{b}
\end{equation}

With the adjusted latent representation $n_{a}$, we can modulate the latent representation $w$ for generating a synthesized image with discrete attributes while retaining other attributes intact. Mathematically, it can be formulated as:

\begin{equation}
    I_{pred}=G_s(w+n_{a}),
\end{equation}

\noindent where $I_{pred}$ represents the synthesized face image with discrete attribute.

\begin{figure}[t]
\begin{center}
   \includegraphics[width=0.85\linewidth]{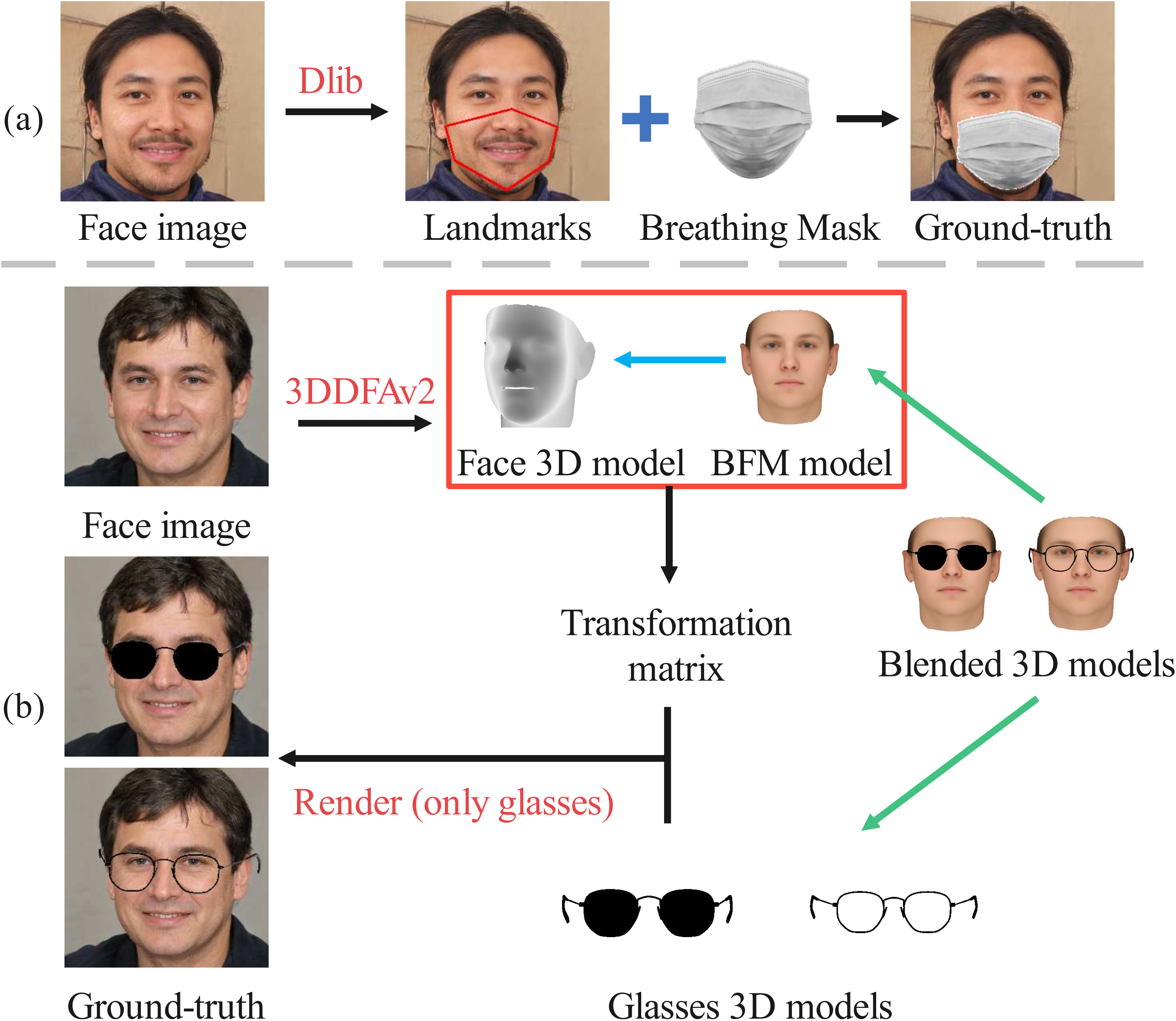}
\end{center}
   \caption{Pipeline of generated synthetic dataset. (a): Ground-truth of face image generation with face mask. (b): Ground-truth of face image generation with different eye glasses.}
\label{fig:meglass}
\end{figure}

\begin{figure*}[!htb]
\begin{center}
   \includegraphics[width=0.85\linewidth]{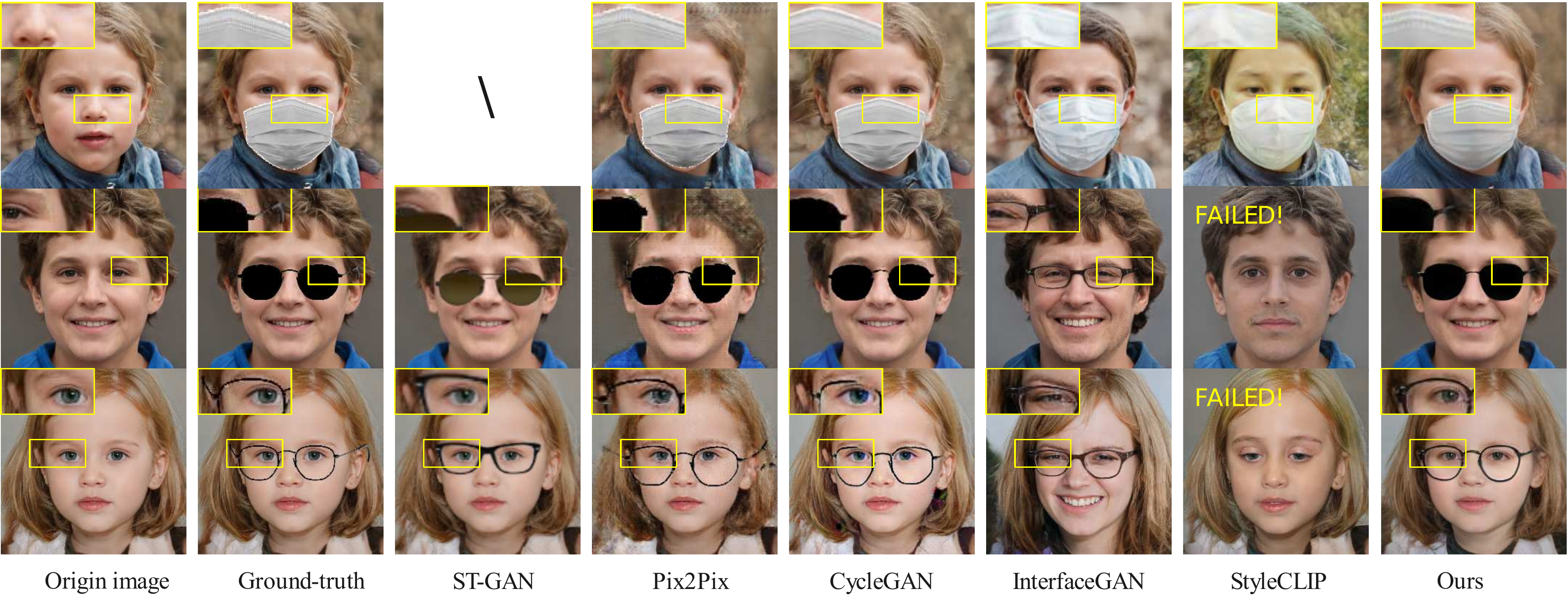}
\end{center}
   \caption{Representative visual results of different methods. Our method outperforms other methods on visual quality.}
\label{fig:refinenet-manipulate}
\end{figure*}

% \begin{figure}[!htb]
% \begin{center}
% %\fbox{\rule{0pt}{2in} \rule{0.9\linewidth}{0pt}}
%   \includegraphics[width=1\linewidth]{img_new/stylegan.eps}
% \end{center}
%   \caption{Diagram of StyleGAN2 training. $G$ and $D$ denote Generator and Discriminator, respectively.}
% \label{fig:stylegan}
% \end{figure} 

\begin{figure}[!h]
\begin{center}
   \includegraphics[width=0.87\linewidth]{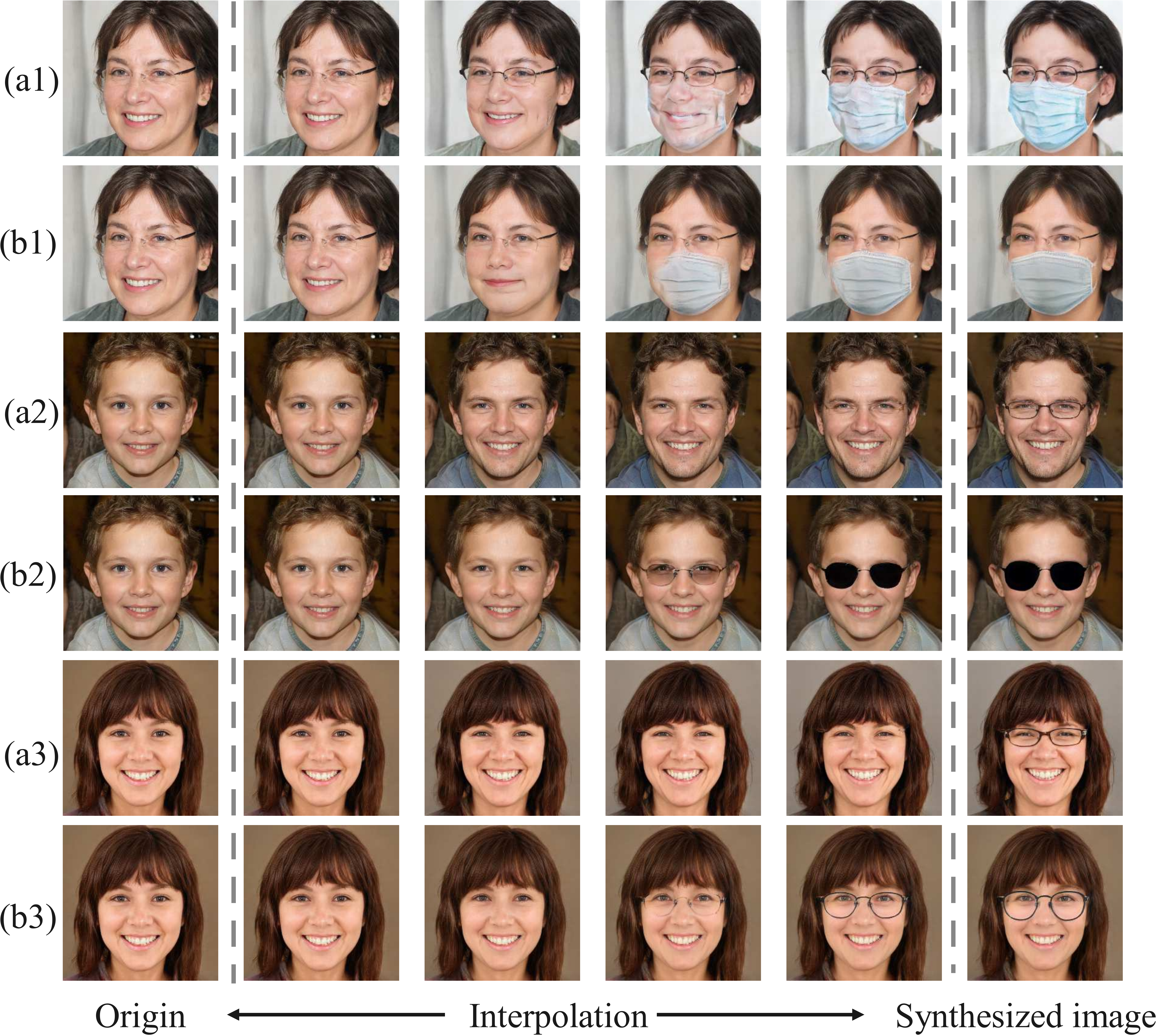}
\end{center}
   \caption{Representative visual results generated by interpolation on latent space of GAN models. (a1-a3): InterfaceGAN, (b1-b3): Our method. Our method outperforms InterfaceGAN in terms of the fluidity and identity preservation.}
\label{fig:interpolation}
\end{figure}

\subsection{Optimization}
\label{sec:optimization}

We adopt MSE (Mean Squared Error) to evaluate the content loss of synthesized image, which can be formulated as:

\begin{equation}
    L_{mse}=||I_{pred}-I_{gt}||_2^2,
\end{equation}
\noindent where $I_{gt}$ represents Ground-truth in Sec.~\ref{sec:synthesic}. We adopt LPIPS~\cite{lpips} loss to evaluate feature-level inconsistency, which can be formulated as:

\begin{equation}
L_{f}=\Phi(I_{pred},I_{gt}),
\end{equation}
\noindent where $\Phi$ represents LPIPS~\cite{lpips} loss.

In order to explicitly guide the network to converge in the direction of face image with discrete attribute, we adopts a class loss which can be formulated as:
\begin{equation}
L_{c}=1-F_{det}(I_{pred}),
\end{equation}

\noindent where $F_{det}$ represents the classification network which is the same as Eq.~\ref{eq:1} of the discrete attribute.

Finally, the total loss for our method is:

\begin{equation}
\begin{split}
    L_{all}= L_{mse}+\lambda_{1} L_{f} +\lambda_{2} L_{c},
\end{split}
\end{equation}

\noindent where $\lambda_{1}$ and $\lambda_{2}$ are empirically defined parameters.

\section{Experiments}
\label{exp:1}
\subsection{Implementation Details}
The backbone of our face image encoder is a pre-trained ArcFace~\cite{arcface}. In our network training, Adam~\cite{adam} optimizer is applied to train our model with $30$ epochs. The batch size is set to $10$. The initial learning rate is $0.01$ and multiplied by $0.8$ after each $5$ epoch. The parameters in the generator are fixed during the whole training process. We evaluate our method on $1,000$ images randomly sampled from the pre-trained StyleGAN2. All experiments are performed on a single GPU (RTX-3090), and PyTorch 1.6.0.

\subsection{Dataset setting}
\subsubsection{Our MEGN for training generator}

It is worth noting that the generator in StyleGAN2~\cite{stylegan2} trained by FFHQ cannot successfully generate images with face mask and eyeglasses. Existing face image datasets with discrete attribute are mainly designed for the task of face detection~\cite{rmfd}, in which the face targets are generally small and fuzzy. Hence, they are not consistent with the distribution of FFHQ. Although there are some synthetic fake datasets~\cite{MaskedFaceNet}, they are not realistic in subjectivity for network training. 

To obtain high-resolution images close to the distribution of FFHQ, we manually construct our MEGN (Face Mask and EyeGlasses images crawled from Google and Naver), which includes $5,000$ face images with the attributes of wearing a face mask and eyeglasses (resolution $256\times256$). All data in this dataset are carefully crawled from Google and Naver, aligned by Dlib~\cite{dlib}. Then, we manually remove the inaccurate and blurred images.

We pre-train the generator $G_s$ StyleGAN2 on a mixed FFHQ and MEGN dataset, enabling the generator to generate informative images with discrete attributes. To the best of our knowledge, our MEGN is the first realistic, high-definition dataset of face images with discrete attributes, especially face mask. Subsequent experiments (as shown in Sec.~\ref{exp:1}) have proved that the complement of MEGN is quite useful for the representation and decoupling of discrete attributes in the latent space.

\subsubsection{Synthetic dataset for training 3D-aware Fusion Network}
\label{sec:synthesic}

Due to the lack of 3D models, we adopt MaskTheFace~\cite{masktheface} to synthesize the face mask image. Specifically, face mask is applied to the face with landmarks detected by Dlib~\cite{dlib}. The procedure of synthetic generation is depicted in Fig.~\ref{fig:meglass}~(a). In glasses image synthesis, it is difficult to precisely locate the feet of glasses due to the ambiguity of depth and self-occlusion, resulting in an unrealistic image. Highly motivated by the works in \cite{meglass,brief, zhang3d, Jie2022Customize,8820128,8440073,zhu1,zhu2,zhu3,zhu4,zhu5,zhu6}, we also adopt 3DDFAv2~\cite{3ddfav2} to obtain 3D face representation~\cite{3dmm} of face image and further find the transformation matrix between BFM~\cite{bfm} and ours. The transformation matrix will be applied to glasses 3D models (pre-registered to BFM). Finally, we render the glasses on top of the current face in Fig.~\ref{fig:meglass}~(b). Although we adopt synthetic images as Ground-truth, extensive experiments reveal that our results exceed Ground-truth in terms of many metrics. Extensive experiments have proven our method is generalisable, and for other discrete attributes, the network can achieve more realistic effect by simply simulating Ground-truth as shown in Fig.~\ref{fig:meglass}.

\begin{table*}[htbp]
\caption{Experimental results on retaining face attributes in synthesizing. Re-score is adopted to evaluate the effect of retaining face attributes. The best Re-score is highlighted in bold. FM denotes face mask, SG denotes sun glasses, and FG denotes frame glasses. The best results are highlighted in red, and the second-best results are highlighted in blue.}
\begin{center}

\begin{tabular}{c|ccc|ccc|ccc|ccc} 
\hline
\multirow{2}{*}{Methods}      & \multicolumn{3}{c|}{Male ($\downarrow$)}                                                                           & \multicolumn{3}{c|}{Young ($\downarrow$)}                                                                          & \multicolumn{3}{c|}{Pose ($\downarrow$)}                                                                           & \multicolumn{3}{c}{Eyeglasses}                                                                           \\ 
\cline{2-13}
                              & FM                               & SG                             & FG                             & FM                               & SG                             & FG                             & FM                               & SG                             & FG                             & FM ($\downarrow$)                            & SG ($\uparrow$)                          & FG ($\uparrow$)                            \\ 
\hline
Ground-truth                  & \textcolor{blue}{\textbf{0.024}} & 0.094                            & \textbf{\textcolor{red}{0.056}}  & \textbf{\textcolor{blue}{0.055}} & 0.131                            & 0.060                            & 0.035                            & \textbf{\textcolor{blue}{0.002}} & 0.014                            & \textcolor{blue}{\textbf{0.364}} & \textbf{\textcolor{blue}{0.992}} & \textcolor{blue}{\textbf{0.987}}   \\
ST-GAN~\cite{stgan}      & -                                & 0.102                            & 0.174                            & -                                & 0.102                            & 0.210                            & -                                & 0.022                            & 0.013                            & -                                & 0.986                            & 0.959                              \\
Pix2Pix~\cite{pix2pix}     & 0.052                            & 0.180                            & 0.149                            & 0.188                            & 0.118                            & 0.101                            & 0.030                            & 0.014                            & \textcolor{red}{\textbf{0.006}}  & 0.603                            & \textcolor{blue}{\textbf{0.992}} & \textbf{\textcolor{blue}{0.987}}   \\
CycleGAN~\cite{cyclegan}     & 0.028                            & \textcolor{blue}{\textbf{0.091}} & \textbf{\textcolor{blue}{0.060}} & \textbf{\textcolor{red}{0.032}}  & \textcolor{blue}{\textbf{0.091}} & \textbf{\textcolor{blue}{0.075}} & \textbf{\textcolor{blue}{0.029}} & 0.007                            & 0.015                            & 0.395                            & \textbf{\textcolor{blue}{0.992}} & \textcolor{blue}{\textbf{0.987}}  \\
InterfaceGAN~\cite{interfacegan} & 0.116                            & 0.163                            & 0.151                            & 0.106                            & 0.352                            & 0.356                            & 0.043                            & 0.016                            & 0.016                            & 0.367                            & 0.958                            & 0.985                              \\
% StyleCLIP~\cite{styleclip} & 0.025                            & 0                            & 0                            & 0.050                            & 0                            & 0                            & 0.014                            & 0                            & 0                            & 0.024                            & 0                            & 0                              \\
Ours                          & \textbf{\textcolor{red}{0.016}}  & \textbf{\textcolor{red}{0.071}}  & 0.127                            & 0.094                            & \textbf{\textcolor{red}{0.014}}  & \textbf{\textcolor{red}{0.026}}  & \textbf{\textcolor{red}{0.019}}  & \textbf{\textcolor{red}{0.000}}  & \textcolor{blue}{\textbf{0.012}} & \textbf{\textcolor{red}{0.191}}  & \textcolor{red}{\textbf{0.993}}  & \textbf{\textcolor{red}{0.988}}    \\
\hline
\end{tabular}
\end{center}
\label{tab:mask-rescore}
\end{table*}

\subsection{Qualitative Experiments}
\subsubsection{Synthesis Face Image with Discrete Attributes}

For the task of face image synthesis, it is important to generate pleasing visual details (such as coherent edges and complete structure) while keeping other attributes unchanged. The qualitative results are shown in Fig.~\ref{fig:refinenet-manipulate}. ST-GAN~\cite{stgan} locates the glasses in the wrong place, which results in an unrealistic appearance. Pix2Pix~\cite{pix2pix} generates noise artifacts while synthesizing. CycleGAN~\cite{cyclegan} generates synthesized images with incoherent edges and incomplete masks. InterfaceGAN~\cite{interfacegan} is capable of synthesizing realistic face images with discrete attributes but fails to retain other face attributes. StyleCLIP~\cite{styleclip} tends to generate inaccurate images. Apparently, our method keeps other face attributes intact as in composition-based methods (Ground-truth, ST-GAN~\cite{stgan}) and image-to-image based methods (Pix2Pix~\cite{pix2pix}, CycleGAN~\cite{cyclegan}), and synthesizes face images with visual details as in semantic-based methods~\cite{interfacegan}. Ground-truth suffers from obvious aliasing and has unsmooth edges. Overall, our method outperforms other methods on visual quality especially anti-aliasing and achieves state-of-the-art results.

\begin{table*}[!htbp]
\caption{User study results for face image synthesis on face mask, sun glasses, and frame glasses. Original feature retention (ORF), mask synthesis comfort (MSC), and overall quality of synthesis (OQS) are three metrics for evaluation.}
\begin{center}
\begin{tabular}{c|ccc|ccc|ccc} 
\hline
\multirow{2}{*}{Methods}      & \multicolumn{3}{c|}{Face Mask}                                                                    & \multicolumn{3}{c|}{Sun Glasses}                                                                  & \multicolumn{3}{c}{Frame Glasses}                                                                  \\ 
\cline{2-10}
                              & OFR ($\uparrow$)                              & MSC ($\uparrow$)                             & OQS ($\uparrow$)                             & OFR ($\uparrow$)                             & MSC ($\uparrow$)                             & OQS ($\uparrow$)                             & OFR ($\uparrow$)                             & MSC ($\uparrow$)                             & OQS ($\uparrow$)                              \\ 
\hline
Ground-truth                  & \textbf{\textcolor{red}{4.092}}  & 2.762                            & 3.062                            & \textbf{\textcolor{blue}{3.977}} & 2.985                            & 3.308                            & \textbf{\textcolor{blue}{4.031}} & 2.869                            & 3.254                             \\
ST-GAN~\cite{stgan}                        & /                                & /                                & /                                & 3.569                            & 2.369                            & 2.554                            & 3.715                            & 2.869                            & 3.038                             \\
Pix2Pix~\cite{pix2pix}      & 3.277                            & 2.685                            & 2.838                            & 2.712                            & 2.185                            & 2.269                            & 3.223                            & 2.438                            & 2.685                             \\
CycleGAN~\cite{cyclegan}     & 3.923                            & 2.838                            & \textcolor{blue}{\textbf{3.162}} & 3.754                            & 2.846                            & 3.100                            & 3.915                            & 2.762                            & 3.192                             \\
InterfaceGAN~\cite{interfacegan} & 2.069  & \textbf{\textcolor{red}{3.985}}  & 3.115                            & 3.408                            & \textcolor{red}{\textbf{4.285}}  & \textbf{\textcolor{blue}{3.823}} & 3.169                            & \textcolor{red}{\textbf{4.277}}  & \textbf{\textcolor{blue}{3.877}}  \\
StyleCLIP~\cite{styleclip} & 3.078                            & 1.674  & 2.154                            & 3.215                            & 1.349  & 2.679 & 3.219                            & 1.561  & 3.476  \\
Ours                          & \textbf{\textcolor{blue}{3.992}} & \textcolor{blue}{\textbf{3.715}} & \textbf{\textcolor{red}{3.854}}  & \textbf{\textcolor{red}{4.077}}  & \textcolor{blue}{\textbf{3.831}} & \textbf{\textcolor{red}{3.947}}  & \textbf{\textcolor{red}{4.077}}  & \textbf{\textcolor{blue}{4.131}} & \textbf{\textcolor{red}{4.138}}   \\
\hline
\end{tabular}
\end{center}
\label{tab:mos}
\end{table*}

\subsubsection{Image Interpolation}
To comprehensively analyze the semantic property, we adopt face image interpolation, which explores the semantic information in face synthesis. According to~\cite{idinvert}, a suitable synthesis should change the face mask gradually while keeping other attributes unchanged. As shown in Fig.~\ref{fig:interpolation}, some representative examples of InterfaceGAN~\cite{interfacegan} implement face synthesis with face mask while suffering from obvious changes in light attribute ($(a_1)$), age attribute ($(a_2)$), and shapes of eyes ($(a_2)$, $(a_3)$). We find that the image interpolation in our method is reasonable in semantics and does not change the attributes which should be retained compared to InterfaceGAN. For example, our method does not distort the structure of the face or change other attributes of the face. In contrast,  both structure distortion and attributes drift of the face occur in the synthesis of InterfaceGAN~\cite{interfacegan}. StyleCLIP [37] is quite hard to modify images with discrete attribute correctly, so it is not listed in this experiment. Apparently, our method achieves remarkable performance in the image interpolation task.

\subsection{Quantitative Experiments}

\subsubsection{Synthesis Performance Evaluating with Re-score}
Here, we adopt Re-score~\cite{interfacegan-pami,higan} to evaluate the ability to retain attributes via predicting the confidence of face attributes fidelity before and after face synthesis. For a fair comparison, we directly borrow the trained prediction models from the official repository in IALS~\cite{ials}. Besides, due to the large coverage proportion of face mask, some concealed attributes, such as smiles, are not considered in our experiments. The detailed results are listed in Tab.~\ref{tab:mask-rescore} and we can observe that our method outperforms other methods on most attributes according to the Re-score metric. Our method outperforms InterfaceGAN~\cite{interfacegan} on the young attribute in frame glasses synthesis by $0.330$ in terms of Re-score. Notably, our method outperforms Pix2Pix~\cite{pix2pix} on the eyeglasses attribute and young attribute in face mask synthesis by $0.412$ and $0.094$ in terms of Re-score, respectively. Generally speaking, image composition based methods (Ground-truth, ST-GAN~\cite{stgan}) and image-to-image based methods (Pix2Pix~\cite{pix2pix}, CycleGAN~\cite{cyclegan}) should achieve optimal results on Re-score because they are directly affixed with a discrete attribute or will learn a fine-grained pixel-level mapping relationship. Apparently, our method is comparable with image composition-based and image-to-image translation methods and even outperforms them in some attributes. Although InterfaceGAN~\cite{interfacegan} achieves promising Re-scores, it greatly changes other attributes, as shown in Fig.~\ref{fig:refinenet-manipulate}. StyleCLIP~\cite{styleclip} edits incorrect images and the resulting images are almost identical to the original, so it is not listed in the table. Overall, our method achieves promising results in terms of Re-score.

\begin{figure*}[!h]
\begin{center}
  \includegraphics[width=0.87\linewidth]{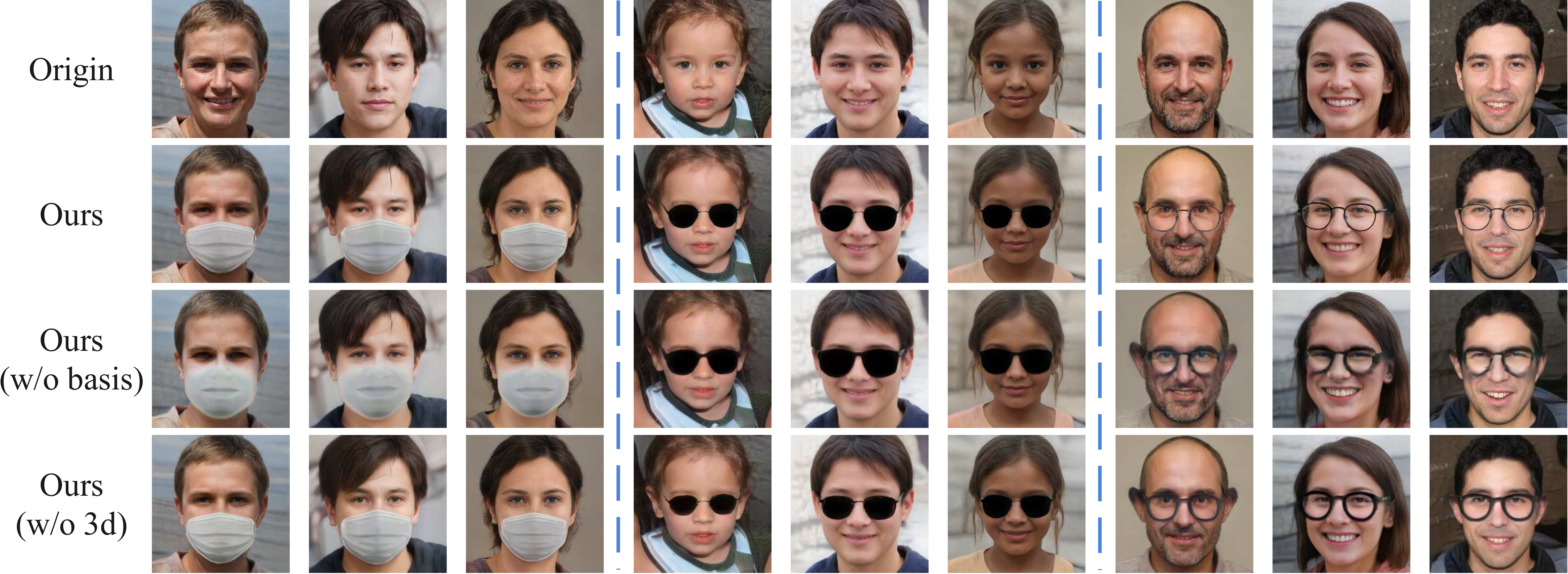}
\end{center}
  \caption{
Ablation study for Ours, Ours (w/o basis) and Ours (w/o 3d). Ours (w/o basis) denotes our method without semantic prior basis and Ours (w/o 3d) denotes our method without 3d information.}
\label{fig:ablation}
\end{figure*}

\begin{table*}[htb]
\label{tab:ablation}
\caption{Experimental results on Re-score for evaluating the importance of semantic prior basis and 3d information. Best scores are highlighted in red.}
\begin{center}
\begin{tabular}{c|ccc|ccc|ccc|ccc} 
\hline
\multirow{2}{*}{Methods} & \multicolumn{3}{c|}{Male ($\downarrow$)}                                                                        & \multicolumn{3}{c|}{Young ($\downarrow$)}                                                                       & \multicolumn{3}{c|}{Pose ($\downarrow$)}                                                                         & \multicolumn{3}{c}{Eyeglasses}                                                                       \\ 
\cline{2-13}
                         & FM                              & SG                            & FG                            & FM                              & SG                            & FG                            & FM                              & SG                            & FG                             & FM ($\downarrow$)                           & SG ($\uparrow$)                         & FG ($\uparrow$)                          \\ 
\hline
Ours                     & \textbf{\textcolor{red}{0.016}} & \textbf{\textcolor{red}{0.071}} & \textbf{\textcolor{red}{0.128}} & \textbf{\textcolor{red}{0.094}} & 0.014                           & \textbf{\textcolor{red}{0.026}} & \textbf{\textcolor{red}{0.019}} & \textbf{\textcolor{red}{0.000}} & \textbf{\textcolor{red}{0.012}}  & 0.191                           & \textbf{\textcolor{red}{0.993}} & \textbf{\textcolor{red}{0.988}}  \\
w/o basis                & 0.096                           & 0.132                           & 0.228                           & 0.204                           & 0.051                           & 0.072                           & \textbf{\textcolor{red}{0.019}} & 0.015                           & 0.038                            & \textbf{\textcolor{red}{0.015}} & 0.992                           & 0.974                            \\
w/o 3d                   & 0.041                           & 0.080                           & 0.214                           & 0.126                           & \textbf{\textcolor{red}{0.004}} & 0.040                           & \textbf{\textcolor{red}{0.019}} & 0.011                           & \textbf{\textcolor{red}{0.012}} & 0.192                           & 0.992                           & 0.987                            \\
\hline
\end{tabular}

\end{center}

\label{tab:ablation}
\end{table*}

\subsubsection{User Study}
User study is a human evaluation metric for verifying the quality of synthesized images~\cite{vtnfp}. To test the quality of generated images comprehensively, we adopt three testing metrics: original feature retention (ORF) for evaluating identity preservation ability while manipulating, mask synthesis comfort (MSC) for evaluating the performance of discrete attribute manipulating, and overall quality of synthesis (OQS) for evaluating the authenticity of global face manipulating. The scores of the above three metrics all range from $1$ to $5$. We invited $200$ volunteers, and each volunteer was randomly given five sets of images randomly selected from $1,000$ groups. Each set includes eight images, i.e., original image, Ground-truth, ST-GAN~\cite{stgan}, Pix2Pix~\cite{pix2pix}, CycleGAN~\cite{cyclegan}, InterfaceGAN~\cite{interfacegan}, StyleCLIP~\cite{styleclip} and Ours. Every volunteer has the duty to score each set of images separately by three metrics. 

The detailed results are listed in Tab.~\ref{tab:mos}. From Tab.~\ref{tab:mos}, we can observe that the image composition-based methods and image-to-image based methods have good ORF but unpromising MSC on discrete masks. InterfaceGAN~\cite{interfacegan} achieves the best MSC but poor ORF. Although the performance of our method on MSC is not the best, it is very close to InterfaceGAN~\cite{interfacegan}. About OFR, our method under-performs Ground-truth only $0.1$ in face mask. But our method apparently outperforms other methods on OFR in sun glasses and frame glasses synthesis. On OQS, our method outperforms InterfaceGAN~\cite{interfacegan} by $0.739$, $0.124$, and $0.261$ on face mask, sun glasses, and frame glasses, respectively. Apparently, our method achieves state-of-the-art performance against other methods.

% \begin{figure*}[!h]
% \begin{center}
% %\fbox{\rule{0pt}{2in} \rule{0.9\linewidth}{0pt}}
%   \includegraphics[width=1\linewidth]{img_new/inversion.eps}
% \end{center}
%   \caption{Representative results for real image synthesis. The 3rd and 4th, 5th and 6th, 7th and 8th columns denote Breathing Mask synthesis, Frame Glasses Mask synthesis and Sun Glasses Mask synthesis respectively.}
% \label{fig:realimg}
% \end{figure*}

% \subsection{Real Image Synthesis}
% To verify the robustness of our method, we conduct our experiments on real images. Since the latent representation of a real image is missing, an inverted way is desiderated. Here, we adopt an encoder-based method to obtain the latent representation. We trained e4e~\cite{e4e} with Restyle~\cite{restyle} on our BGGN and FFHQ~\cite{stylegan} for inverting real images to a latent representation. The synthesized image of the latent representation would be quite similar to the original one. The representative results are shown in Fig.~\ref{fig:realimg}. From Fig.~\ref{fig:realimg}, we can see that the details of the original images are well retained, and the wearing areas of the masks are generated clearly. While InterfaceGAN is failed to retain the details and synthesize blurry images.

\subsection{Ablation Study}
\label{exp:ablation}
To verify the effectiveness of our semantic prior basis and 3D-aware fusion network, we conduct two ablation experiments both qualitatively and quantitatively, as shown in Fig.~\ref{fig:ablation} and Tab.~\ref{tab:ablation}. 

Comparing the results without semantic prior basis and without 3D information, we observe that synthesized images generated via semantic prior basis are more visually realistic than the results without semantic prior basis (denoted by "Ours (w/o basis)") and the results without 3D information (denoted by "Ours (w/o 3d)"). The results generated via semantic prior basis and 3D information have coherent synthesizing edges and intact face information, while the results without the semantic prior basis are not photo-realistic. This may be induced by the large area of the face mask compared with relatively small face images, making the network challenging to learn discrete semantic attribute representations of GAN directly. In sun glasses and frame glasses synthesis, the shape of glasses is quite blurry and aliasing without the help of semantic prior basis. In some details (such as lips and hair), the results without 3D information tend to be changed lightly. The lack of 3D information embedding makes it difficult for the network to retain the detailed information of the original image. While in our method, the synthesized images with semantic prior basis and 3D-aware embedding are more anti-aliasing and authentic.

In addition, we analyze the quantitative results of our method in terms of Re-score with and without semantic prior basis and 3d information. The detailed results are listed in Tab.~\ref{tab:ablation}. From Tab.~\ref{tab:ablation}, we observe that "Ours" apparently outperforms "Ours (w/o basis)" and "Ours (w/o 3d)" nearly on all attribute metrics. In particular, "Ours" outperforms "Ours (w/o basis)" by $0.061$ on the Male attribute and outperforms "Ours (w/o basis)" by $0.015$ on the pose attribute when synthesizing sun glasses. In addition, the results of "Ours (w/o basis)" are always worse in metrics. Overall, our method can coherently improve the quality of synthesized images qualitatively and quantitatively, especially when semantic prior basis is well relevant to the optimal semantic representation of the GAN model. 

\section{Conclusion}
In this paper, we propose an innovative framework by decomposing semantic discrete attributes representation of GAN into semantic prior basis and offset latent representation. The semantic prior basis will be learned by the SVM classifier in the latent space of GAN and a novel semantic fusion network is proposed to generate offset latent representation of facial attributes with the guidance of face 3D information. In this way, our method can well learn accurate discrete attributes in the facial representation for synthesizing photo-realistic face images. Extensive experiments demonstrate that our method can synthesize photo-realistic face images with discrete attributes while stabilizing other attributes. In the future, we will continue to study the properties of semantics in the latent space of GANs for generic real image editing tasks.

\begin{acks}
This research was supported by the National Key Research and Development Program of China (2020AAA09701), National Science Fund for Distinguished Young Scholars (62125601), National Natural Science Foundation of China (62076024, 62172035, 62006018, 61806017).
\end{acks}

\bibliographystyle{ACM-Reference-Format}
\bibliography{sample-base}

\appendix

\begin{figure*}[!t]
% \begin{center}
  \includegraphics[width=1\linewidth]{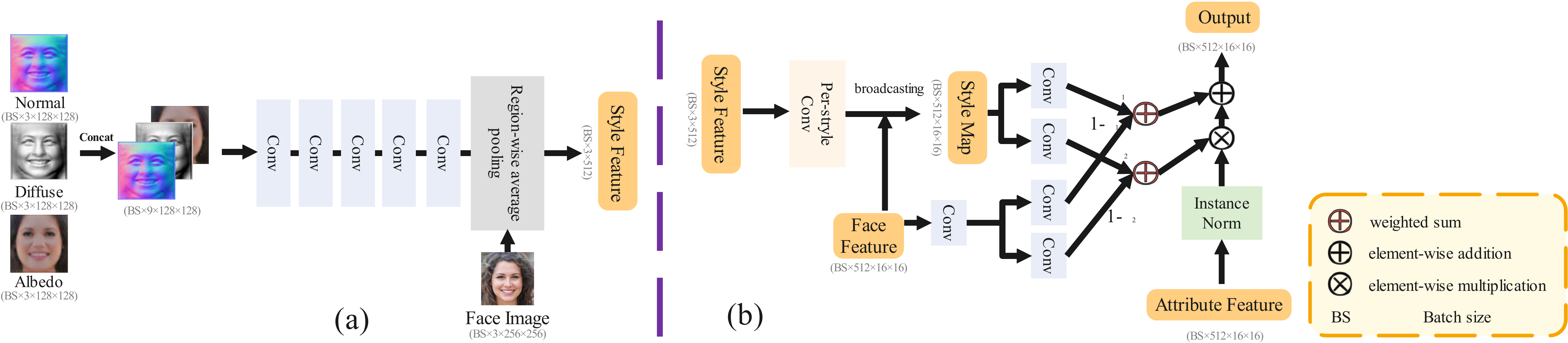}
% \end{center}
\caption{The architecture of style encoder and fusion module.}
\label{fig:arch}
\end{figure*}

\section{Overview}
This supplementary material contains the following parts:

\begin{itemize}
	\item We introduce the architecture details of style encoder and fusion modules of the proposed 3D-aware semantic fusion network.

    \item We provide extra experiments to prove the performance of our method.

    \item We provide some example images in our proposed MEGN (Face Mask and Eyeglasses images crawled from Google and Naver).
    
    \item We propose a video to show discrete attribute manipulation results (see MM.mp4 in the supplement zip).
\end{itemize}

\section{Architecture Details}
Following the networks in SPADE~\cite{spade} and SEAN~\cite{sean}, we design the structure of style encoder and fusion module to synthesize face with discrete attributes.

\subsection{Style Encoder}
As shown in Fig.~\ref{fig:arch} (a), our style encoder consists of  multi-layer convolutions to extract 3D features from the concatenation of face normal map, diffuse map, and albedo images. The region-wise average pooling is adopted here to get style feature. The pooling mainly aims to adapt the dimensions of the subsequent module.

\subsection{Fusion Module}
As shown in Fig.~\ref{fig:arch} (b), our fusion module adopts weighted learning with coefficients to fuse the extracted features. Specifically, the style feature undergoes a per-style convolution and is then broadcast to face feature. In this way, the style map is yielded. The style map is processed by convolution layers to produce pixel normalization values of 3D information. The face feature passes through a convolution layer and then two separate convolution layers to obtain pixel normalization values of face. The learnable weight parameters $\alpha_1$ and $\alpha_2$ during training would adjust the proportion of each variable when fusing with attribute feature.

\section{Extra Experiments}
\begin{table*}[htbp]
\caption{Experimental results of face image quality estimation with SDD-FIQA and SER-FIQ. The best SDD-FIQA and SER-FIQ are highlighted in red, and the second-best results are highlighted in blue. FM denotes face mask, SG denotes sun glasses and FG denotes frame glasses.}
\begin{center}
\begin{tabular}{c|ccc|ccc} 
\hline
\multirow{2}{*}{Methods}      & \multicolumn{3}{c|}{SDD-FIQA~\cite{sddfiqa} ($\uparrow$)}                                                         & \multicolumn{3}{c}{SER-FIQ~\cite{serfiq} ($\uparrow$)}                                                         \\ 
\cline{2-7}
                              & FM                                & SG                              & FG                              & FM                               & SG                             & FG                              \\ 
\hline
Ground-truth                  & 58.017                            & 55.854                            & 66.045                            & 0.874                            & 0.767                            & \textcolor{blue}{\textbf{0.885}}  \\
ST-GAN~\cite{stgan} & /                                 & 53.329                            & 65.691                            & /                                & 0.816                            & 0.875                             \\
Pix2Pix\cite{pix2pix}      & 56.102                            & 50.326                            & 63.319                            & \textbf{\textcolor{blue}{0.880}} & 0.491                            & 0.879                             \\
CycleGAN\cite{cyclegan}     & \textbf{\textcolor{blue}{58.222}} & 55.482                            & \textbf{\textcolor{blue}{66.272}} & 0.878                            & 0.775                            & 0.883                             \\
InterfaceGAN~\cite{interfacegan} & 55.523                            & \textcolor{red}{\textbf{68.652}}  & \textcolor{red}{\textbf{68.507}}  & 0.877                            & \textbf{\textcolor{red}{0.889}}  & \textcolor{red}{\textbf{0.889}}   \\
StyleCLIP~\cite{styleclip} & 55.720                            & 54.694  & 55.574  & \textcolor{blue}{\textbf{0.880}}                            & \textbf{\textcolor{blue}{0.886}}  & \textcolor{blue}{\textbf{0.885}}   \\
Ours                          & \textbf{\textcolor{red}{61.473}}  & \textcolor{blue}{\textbf{56.350}} & 64.789                            & \textcolor{red}{\textbf{0.881}}  & 0.859 & 0.884                             \\
\hline
\end{tabular}
\end{center}
\label{tab:quality}
\end{table*}

\subsection{Image Quality Estimation}
In our experiments, we adopt SDD-FIQA~\cite{sddfiqa} and SER-FIQ~\cite{serfiq} to evaluate the realism of synthesized images. SDD-FIQA~\cite{sddfiqa} and SER-FIQ~\cite{serfiq} are two popular metrics in evaluating the effectiveness of image data for the face recognition task. The higher scores of SDD-FIQA and SER-FIQ denote the better quality of synthesized images. The detailed results are listed in Tab.~\ref{tab:quality}.

From Tab.~\ref{tab:quality}, we observe that our method outperforms other methods in terms of SDD-FIQA and SER-FIQ in face mask synthesis. Notably, our method outperforms InterfaceGAN~\cite{interfacegan} in face mask synthesis by $5.95$ in terms of SDD-FIQA. Although InterfaceGAN achieves a higher SDD-FIQA score ($68.652$) than Ours ($56.350$)  on sum glasses synthesis, it significantly modifies other attributes, as shown in Fig.~\ref{fig:mask},\ref{fig:frame},\ref{fig:sun}. And this is detrimental for many tasks, such as data augmentation for face recognition. Overall, our method achieves promising quality for face image synthesis and can significantly benefit data augmentation in face recognition-related tasks.

\begin{figure}[!t]
% \begin{center}
  \includegraphics[width=1\linewidth]{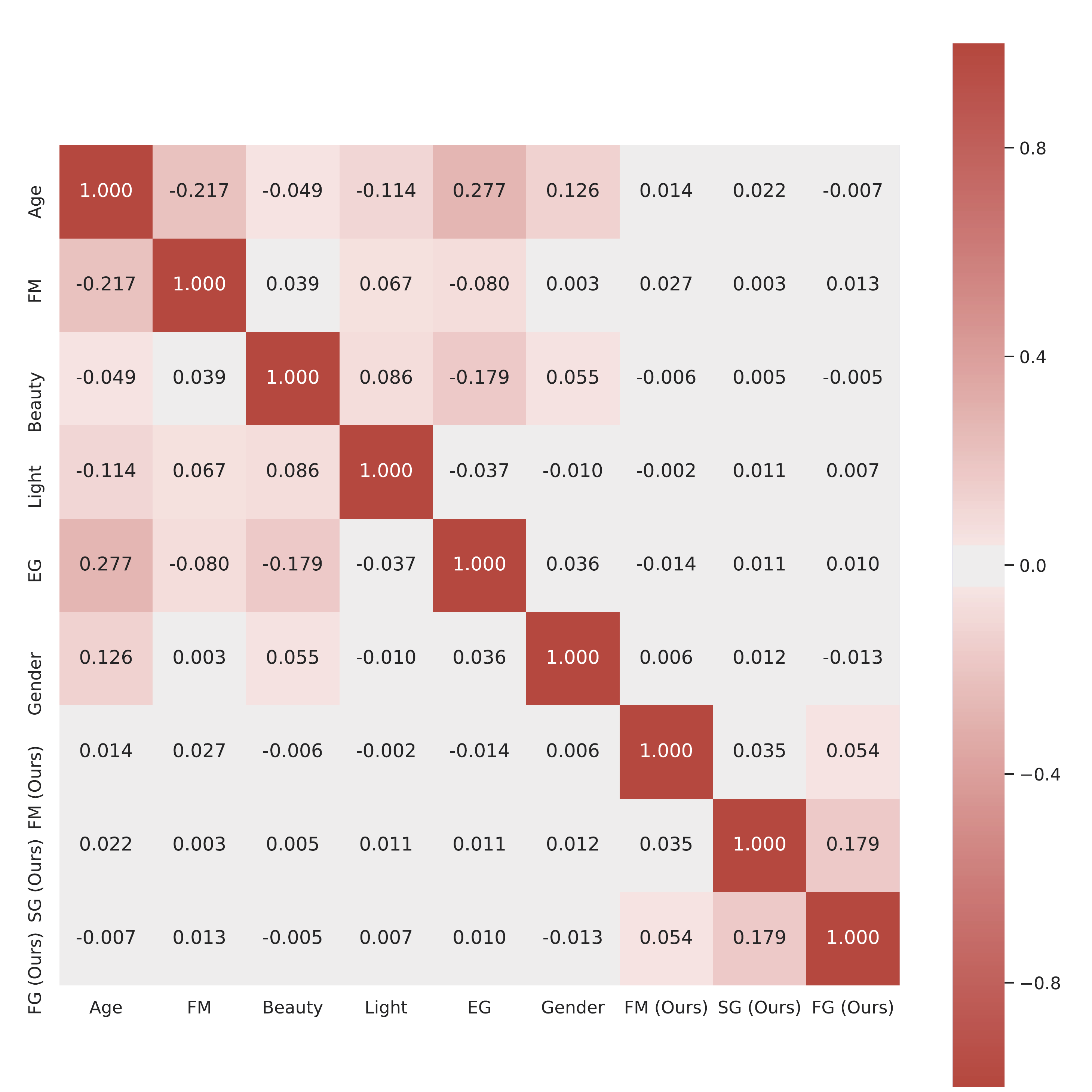}
% \end{center}
\caption{Experimental results of decoupled relationships between attributes. A large value of cosine similarity indicates a bad decoupled degree of two attributes. Our method outperforms InterfaceGAN with
a significant margin.}
\label{fig:cos}
\end{figure}

\subsection{Decoupled Degree Between Attributes}

In this section, we study the decoupled degrees between attributes to reflect if attribute subspace is correctly divided in latent space. Here, we use cosine similarity to measure the decoupled degree between two semantic representations. A large value of cosine similarity indicates a bad decoupled degree of two attributes. We compare our method with InterfaceGAN~\cite{interfacegan}. The detailed results are shown in Fig.~\ref{fig:cos}. The attributes (age, beauty, light, gender, face mask, and glasses) obtained by InterfaceGAN~\cite{interfacegan} have a bad decouple relation between each other. Especially, the decoupled degree between age and face mask reaches $0.217$, the decoupled degree between age and glasses reaches $0.277$, the decoupled degree between beauty and glasses reaches $0.179$. In our method, the decoupled degrees between attributes are all approximate to $0$. This indicates that attribute representations in our method are almost orthogonal with each other. In particular, our method uncouples age with glasses, age with face mask, and beauty with sun glasses, resulting in superior decoupled degree $0.007$, $0.014$, and $0.005$, respectively. Apparently, our method outperforms InterfaceGAN~\cite{interfacegan} in the capability of decoupling relationships between different attributes with a significant margin.

\subsection{Additional Results}
\begin{figure*}[!t]
% \begin{center}
  \includegraphics[width=1\linewidth]{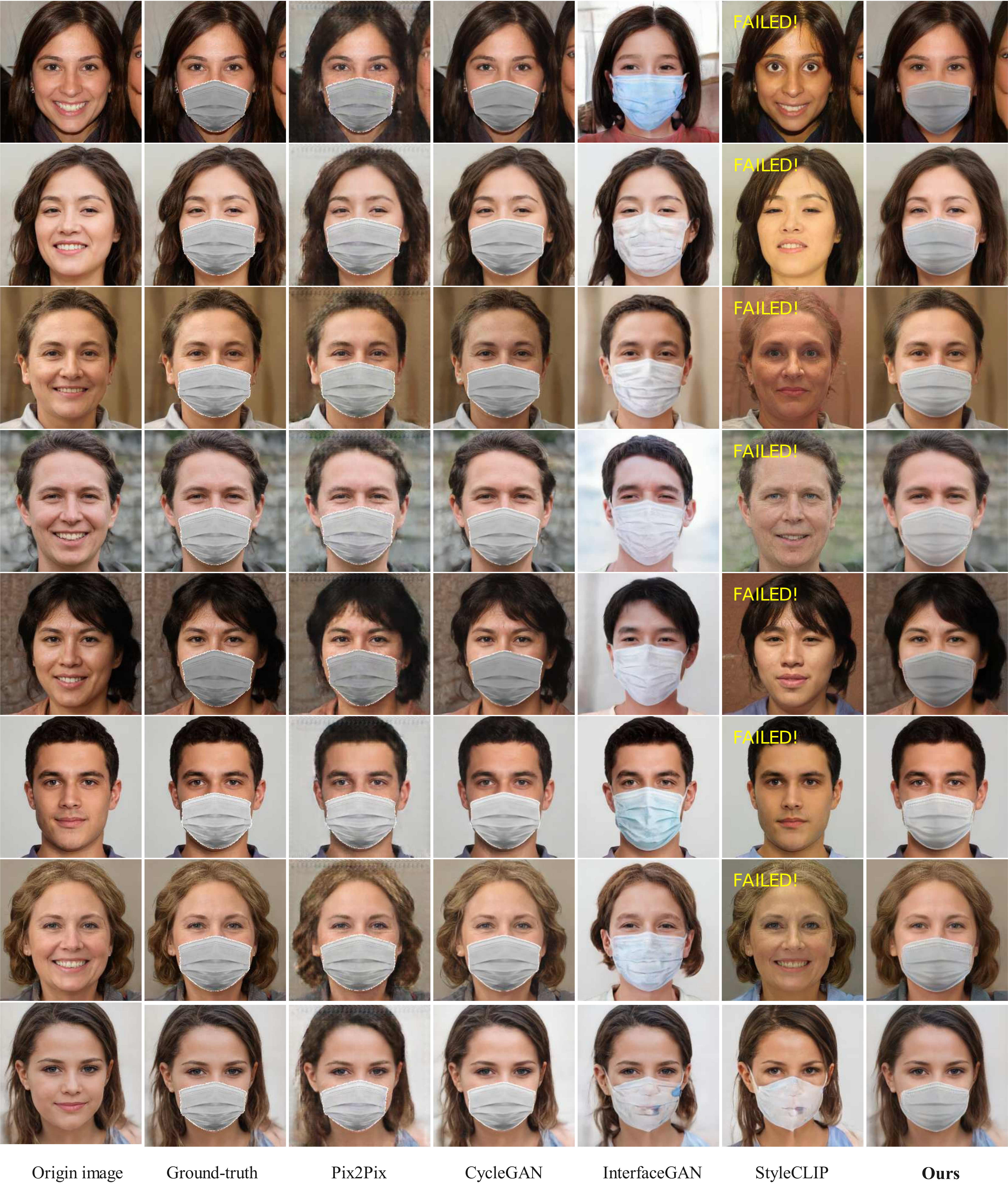}
% \end{center}
\caption{Representative visual results of discrete face mask attribute synthesis. Other methods have poor performance on discrete face mask and fail retain face attributes while editing. Our method outperforms other methods on visual quality especially anti-aliasing and other face attributes intactness.}

\label{fig:mask}
\end{figure*}

\begin{figure*}[!t]
% \begin{center}
  \includegraphics[width=1\linewidth]{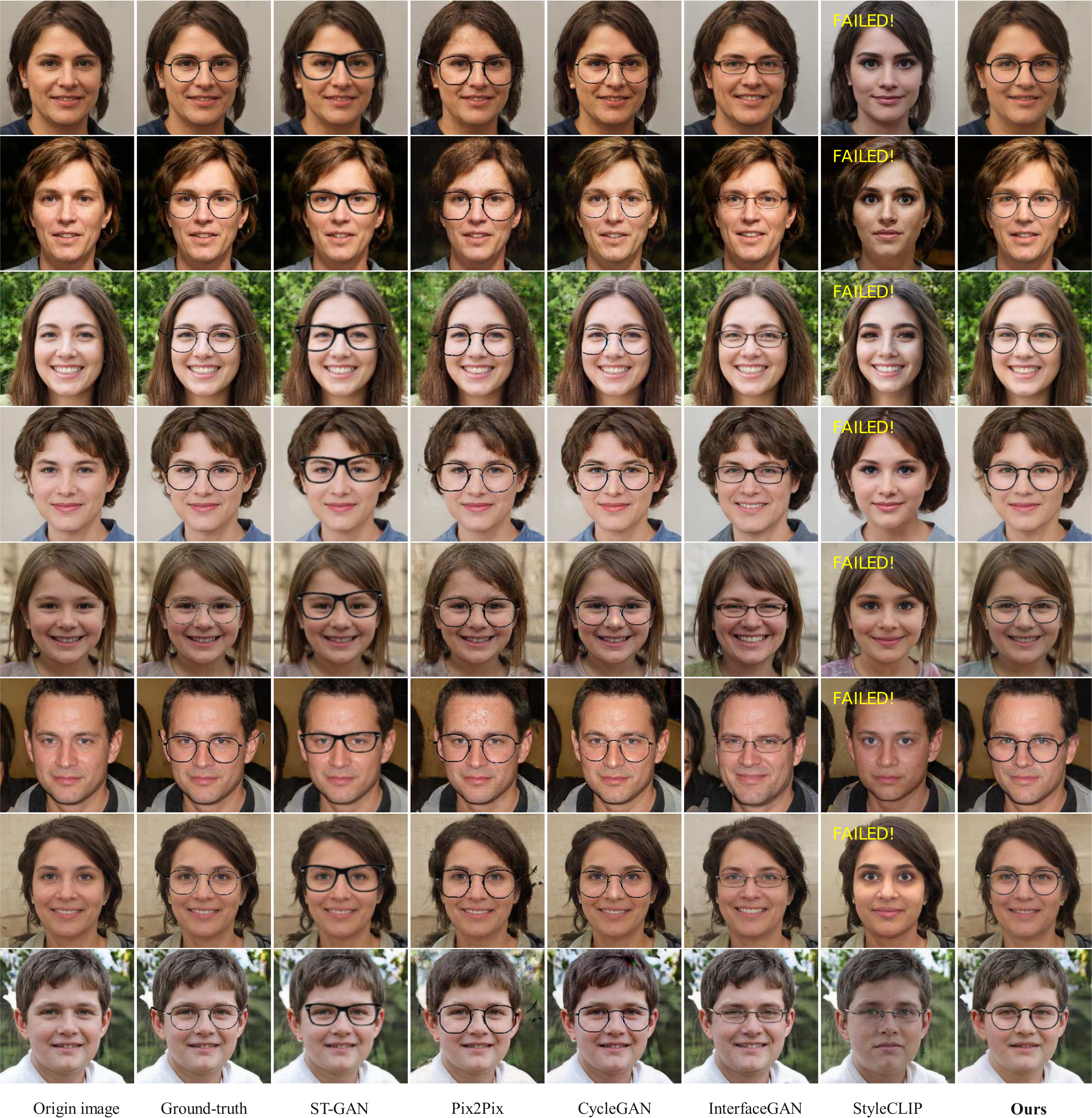}
% \end{center}
\caption{Representative visual results of discrete frame glasses attribute synthesis. Other methods have poor performance on discrete frame glasses and fail retain face attributes while editing. Our method outperforms other methods on visual quality especially anti-aliasing and other face attributes intactness.}
\label{fig:frame}
\end{figure*}

\begin{figure*}[!t]
% \begin{center}
  \includegraphics[width=1\linewidth]{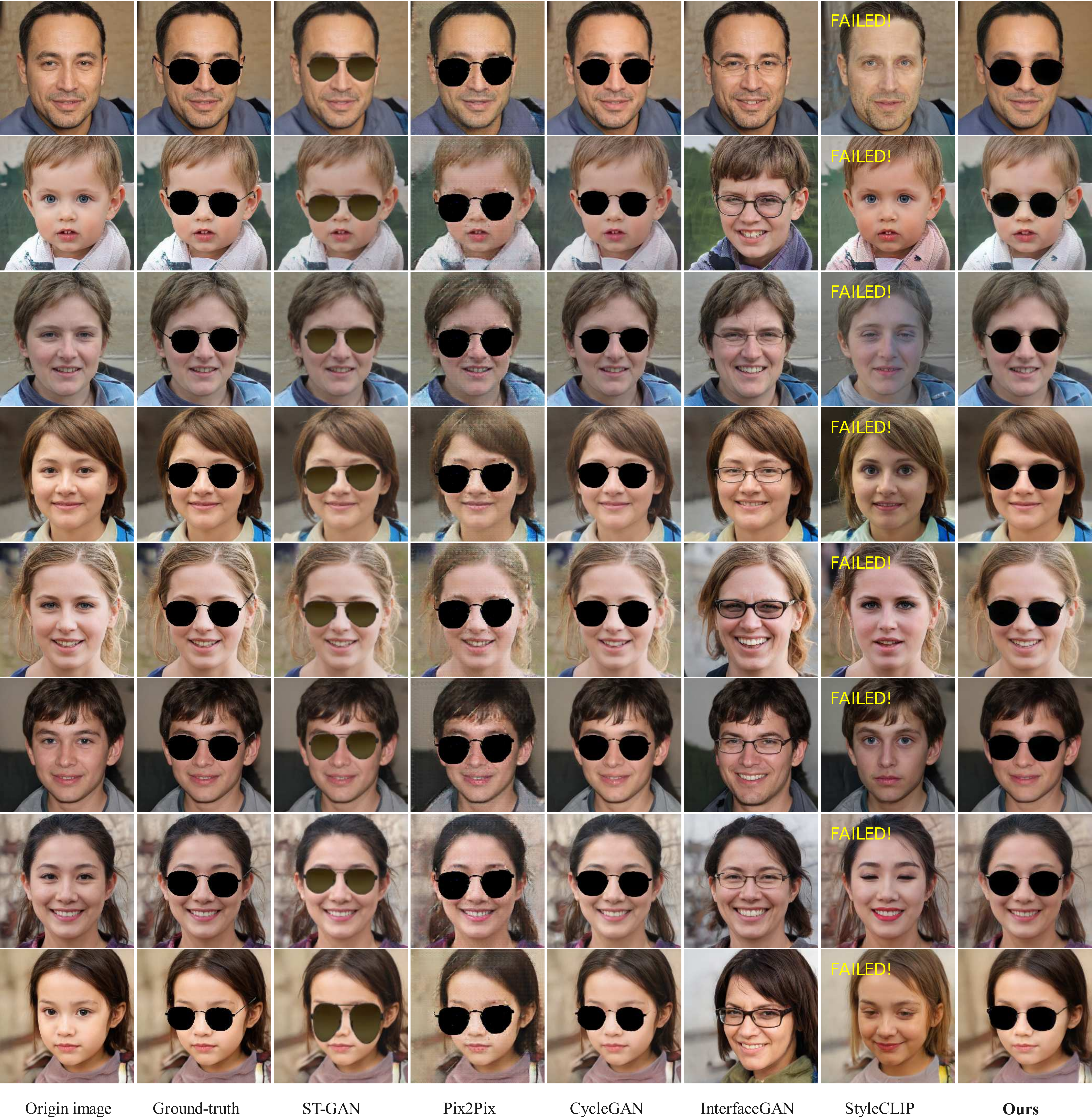}
% \end{center}
\caption{Representative visual results of discrete sun glasses attribute synthesis. Other methods have poor performance on discrete sun glasses and fail retain face attributes while editing. Our method outperforms other methods on visual quality especially anti-aliasing and other face attributes intactness.}
\label{fig:sun}
\end{figure*}

We provide additional results to those presented in the paper. In Fig.~\ref{fig:mask},\ref{fig:frame},\ref{fig:sun}, we show a large number of visual results of face mask, frame glasses and sun glasses synthesis methods separately. Our method keeps other face attributes intact and also synthesizes face images with visual details especially on anti-aliasing.

\section{Our proposed MEGN}
\begin{figure*}[!t]
% \begin{center}
  \includegraphics[width=1\linewidth]{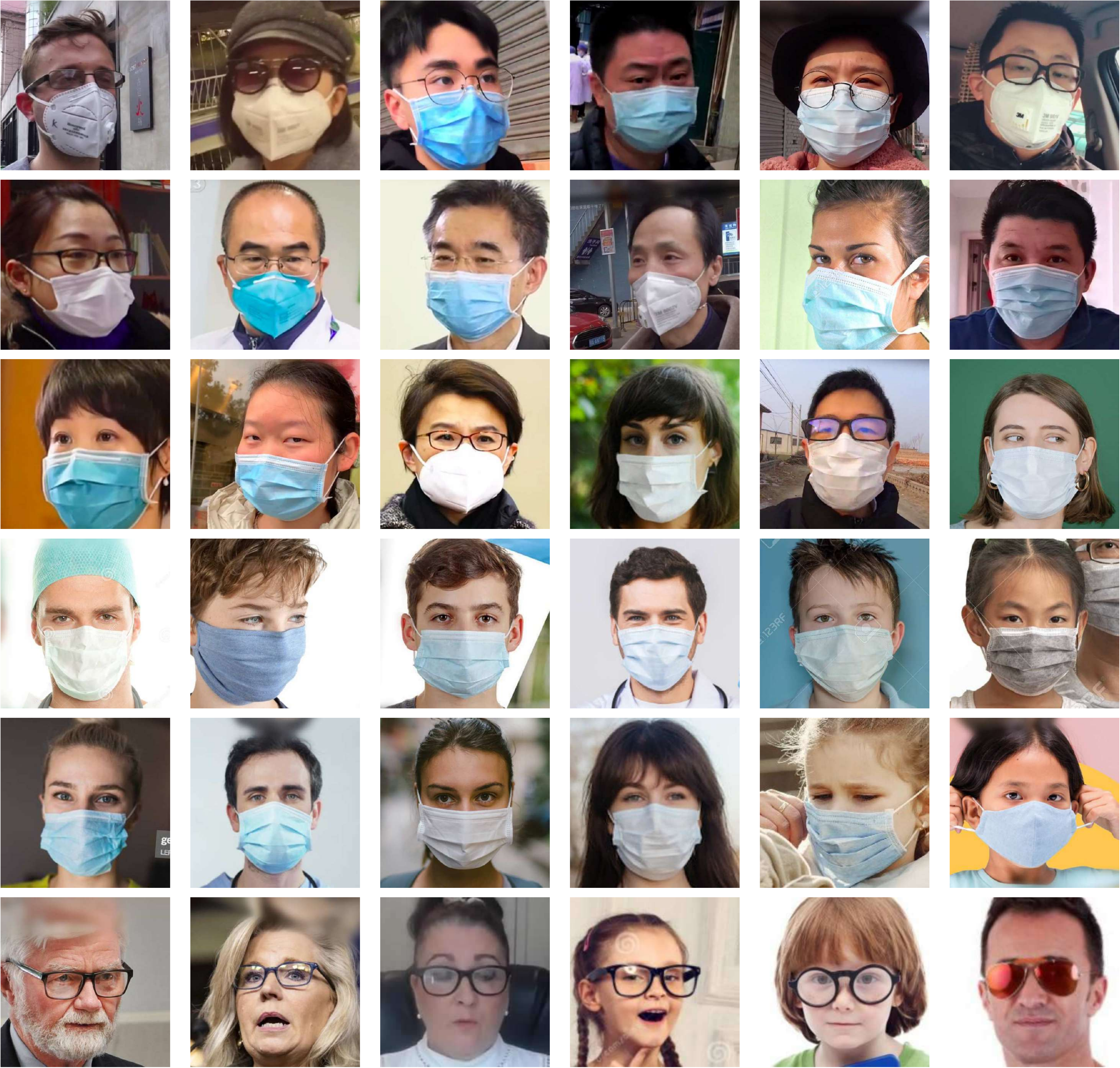}
% \end{center}
\caption{Some example images in MEGN. Our MEGN is the first realistic, high-definition dataset of face images with discrete attributes, especially face mask.}
\label{fig:megn}
\end{figure*}

We propose a big dataset MEGN (Face Mask and Eyeglasses images crawled from Google and Naver) which includes $5,000$ face images with discrete attributes. See Fig.~\ref{fig:megn} for some representative images. Existing face image datasets with discrete attributes only have small and fuzzy images. To the best of our knowledge, our MEGN is the first realistic, high-definition dataset of face images with discrete attributes, especially face mask.

\end{document}